# Thutmose Tagger: Single-pass neural model for Inverse Text Normalization

*Alexandra Antonova, Evelina Bakhturina, Boris Ginsburg*

NVIDIA

{aleksandraa, ebakhturina, bginsburg}@nvidia.com

## Abstract

Inverse text normalization (ITN) is an essential post-processing step in automatic speech recognition (ASR). It converts numbers, dates, abbreviations, and other semiotic classes from the spoken form generated by ASR to their written forms. One can consider ITN as a Machine Translation task and use neural sequence-to-sequence models to solve it. Unfortunately, such neural models are prone to hallucinations that could lead to unacceptable errors. To mitigate this issue, we propose a single-pass token classifier model that regards ITN as a tagging task. The model assigns a replacement fragment to every input token or marks it for deletion or copying without changes. We present a method of dataset preparation, based on granular alignment of ITN examples. The proposed model is less prone to hallucination errors. The model is trained on the Google Text Normalization dataset and achieves state-of-the-art sentence accuracy on both English and Russian test sets. One-to-one correspondence between tags and input words improves the interpretability of the model's predictions, simplifies debugging, and allows for post-processing corrections. The model is simpler than sequence-to-sequence models and easier to optimize in production settings. The model and the code to prepare the dataset is published as part of NeMo project[1].

## 1. Introduction

Inverse text normalization (ITN) is an important post-processing step within an automatic speech recognition (ASR) system. ITN transforms spoken-domain text into its written form. For example, the input expression "on may third we paid one hundred and twenty three dollars" should be converted to "on may 3 we paid $123". The commonly recognized problem is that any automatic conversion can introduce unrecoverable errors that change the meaning of the input. For example, the conversions to "on may 03 we paid 123$" or "on may 03 we paid 123 dollars" are also acceptable since they keep the original meaning, while "on may 30 we paid $123" or "on may 3 we paid $1203" are incorrect. There exist several approaches to ITN:

1. The traditional rule-based approach, based on Weighted Finite-State Transducers (WFST) or regular expressions, provides complete control over the generated output. However, this approach requires linguistic knowledge and is hard to create and maintain to cover all possible cases. Additionally, the rule-based systems usually do not take context into account, which could deteriorate the normalization accuracy.

2. Neural network (NN) based approaches, for example, seq2seq architectures in [1, 2, 3] use a two-step approach. First, a tagger identifies the spans for conversion, and then a decoder translates these spans from spoken to the written domain. A decoder with a *copy* mechanism is used in [4]. NN-based models take context into account and generalize better compared to rule-based systems. However, seq2seq models are prone to hallucinations, and their errors are hard to debug and correct.

3. Hybrid models combine neural seq2seq with WFST rules [5]. The WFST constrains the predictions of the NN model when the NN has low confidence in the prediction or corrects common mistakes in the output of the seq2seq model.

This paper proposes a model, Thutmose Tagger, that treats the ITN task as a tagging problem (Section 2). The goal of tagging is to assign a tag to each input word in the spoken domain sentence so that the concatenation of these tags yields the desired written domain sentence (Figure 1). Our model is NN-based, but has simpler architecture than a seq2seq model: it is a single-pass token classifier.

Our approach is inspired by LaserTagger [6]. LaserTagger shows that many monotonic sequence-to-sequence transformation tasks, such as text simplification or grammar correction, can be reformulated as tagging tasks. It classifies all input words as *keep*, *delete*, or *replacement* tags. The authors propose to collect a vocabulary of replacements tags from the input corpus - exclude all longest common sub-sequences, and regard all non-common fragments as deletions or replacements. The LaserTagger method is not directly applicable to ITN because it can only regard the whole non-common fragment as a single replacement tag, whereas spoken-to-written conversion, e.g. a date, needs to be aligned on a more granular level. Otherwise, the tag vocabulary should include all possible numbers, dates etc. which is impossible. For example, given an example pair "over four hundred thousand fish" - "over 400,000 fish", LaserTagger will need a single replacement "400,000" in the tag vocabulary.

To overcome this problem, we collect the replacement vocabulary based on automatic alignment of spoken-domain words to small fragments of written-domain text along with <SELF> and <DELETE> tags (Section 2.2).

A tagging approach for ITN is also introduced in [7], where each tag represents a sequence of actions needed to convert an input fragment to the written form. However, the authors do not provide an actual implementation of the model and tools to build the training dataset. In Proteno[8], the authors apply a tagging approach to written-to-spoken text normalization (TN) to limit the number of possible decoder outputs. The Proteno approach classifies written tokens into classes and performs written-to-spoken form conversion using manually predefined or automatically learned transformations. Our work considers a reverse problem of spoken to written conversion and relies on an automatic alignment procedure.

The contributions of our paper are the following:

- We present a new model for ITN, which shows state-of-the-art sentence accuracy for English and reduces WER by 3% on Russian hard examples.

---

[1]https://github.com/NVIDIA/NeMo

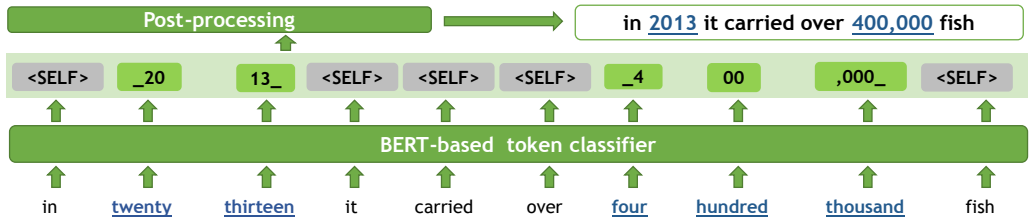

Figure 1: *ITN as tagging: inference example. The sequence of input words is processed by BERT-based token classifier, giving the output tag sequence. Simple rule-based deterministic post-processing gives the final output.*

- We apply an alignment technique from statistical machine translation to automatically find alignment between spoken and written parts of the examples in ITN dataset.
- The model and scripts for dataset construction are open source.[2]

Section 2 describes the models and methods used to build the training dataset. Section 3 reports the experimental results for English and Russian languages and provides the error analysis.

## 2. Proposed approach

The Thutmose Tagger is a neural token classification model. The input is the spoken domain sentence without punctuation. The model outputs tags for all input words. Thutmose is a pretrained BERT encoder [9] with a multi-layer perceptron (MLP) on top, followed by the softmax layer. Fig 1 illustrates how the sentence "in twenty thirteen it carried over four hundred thousand fish" is converted to "in 2013 it carried over 400,000 fish".

### 2.1. Initial data

To train a neural tagging model, we first build aligned datasets for Russian and English languages from the Google Text Normalization (GTN) dataset [10]. The GTN dataset consists of unnormalized (i.e. written form) and normalized (i.e. spoken form) sentence pairs that are aligned *on a phrase-level*. We need to align them on a more *granular level* to get a monotonic one-to-one correspondence between *each spoken word* and corresponding fragments in written form (see Table 1). The goal is to get a restricted vocabulary of target fragments (tags) that can cover most spoken-written pair conversions.

### 2.2. Alignment

We extract all corresponding phrases from GTN Dataset [3] to create a parallel corpus for each semiotic class. We use the Giza++ [11] package to align the resulting parallel corpora. To do the alignment we need to tokenize data first. The spoken text is tokenized by word boundary, while the written part is tokenized as follows: 1) All alphabetic sequences are separate tokens, 2) In numeric sequences each character is a separate token. 3) All non-alphanumeric characters are separate tokens. Additionally, we add an underscore symbol to mark the beginning and end of a sequence for future detokenization. For example, "jan 30,2005" is tokenized as "_jan_ _3 0 , 2 0 0 5_".

We run Giza++ with default settings and join together character-tokens in the written part that are aligned to the same spoken input word. If a spoken input word aligns to nothing, we add a "<DELETE>" tag. As a result we should get a one-to-one alignment for each phrase (see Table 1). There are cases, for which Giza++ alignments are not perfect, for example, consecutive equal digits are sometimes aligned incorrectly. We apply a set of simple regular expressions to correct some systematic incorrect splits (eg. "_1 4000_ => _14 000_", "_1 5,000_ => _15 ,000_").

[2] https://github.com/NVIDIA/NeMo/blob/stable/tutorials/text_processing/ITN_with_Thutmose_Tagger.ipynb
[3] we reserve the same files for testing as in [10]

Table 1: *Examples of alignment. Each section consists of two rows: first contains words from the spoken-domain part, the second row consists of written-domain fragments, aligned to the input words one-to-one.*

| one | thousand | two | hundred | megawatts |
|-----|----------|-----|---------|-----------|
| _1  | <DEL>    | 2   | 00_     | _mw_      |
| н   | e        | й   | ч       | e         | p |
| _n  | a        | <DEL>| t      | u         | re_ |
| r   | and      | b   |         |           |
| _r_ | &        | _b_ |         |           |
| one | hundred  | thousand | dollars |       |
| _1  | 00       | ,000_ | _$<<_  |           |
| fourteen | and  | a   | half    |           |
| _14 | <DEL>    | <DEL>| $\frac{1}{2}$ |     |

### 2.3. Non-Monotonic alignments

One important restriction of the Tagger model is that spoken and written pairs are assumed to be monotonically aligned. Most of the spoken-written pairs in GTN satisfy this requirement. We can detect non-monotonic examples: they have different order of the aligned tokens in Giza++ output. Some non-monotonic cases like special date formats (e.g. "the sixteenth of june two thousand four" - "2004-06-16") have semantically equivalent monotonic versions in the corpus (e.g. "16 June 2004"), so non-monotonic examples can be dropped. We discard all non-monotonic examples, except several common cases for which we encode the information about token movements in the tag itself. This is done before the alignment. Specifically, we move target tokens so that the sequence becomes monotonic and encode the information about the movement with one or two angle brackets, reflecting the direction and type of movement.

Table 2: *Encoding movement information in tags*

| **Input:** | ten square kilometers | ten thousand dollars |
|------------|----------------------|----------------------|
| **Tags:**  | _10_ _²_> _km_       | _10 ,000_ _$<<_      |
| **Swapped:** | _10_ _km_ _²_      | $ _10 ,000_          |

During the detokenization step, we look at the tags with angle brackets and move the corresponding target token to its appropriate place (Table 2).

### 2.4. Tag Vocabulary and Training Dataset

The alignment procedure splits written sentences into fragments and aligns them one-to-one to the spoken form. Then, we count the frequencies of such written fragments and include some predefined number of the most frequent of them in the tag vocabulary. We use 2127 tags for English and 3201 tags for Russian replacement vocabularies. "<SELF>" and "<DELETE>" tags are also added to the tag vocabularies to signify that an input token should be either copied or deleted.

We discard examples where the target part is not fully covered with the tag vocabulary. It is acceptable because most rare fragments result in alignment errors. Nevertheless, the model may find a way to translate any input sequence with its available tag vocabulary at inference time. The exceptions may exist, but they are scarce.

The tagger model sees the whole sentence to make context-dependent decisions. All input tokens outside ITN spans are mapped to "<SELF>" tags during dataset creation. The tagger itself knows nothing about ITN spans, and it simply learns to predict tags for all tokens in a sentence in a single pass.

Both English and Russian training corpora consist of about 2 million sentences, from which 1.5 million are random sentences from the training part of GTN Dataset. Additionally, we sample 500,000 sentences that contain examples of different tags from the tag vocabulary since some tags are really rare.

### 2.5. Post-processing

To get the final output we apply a simple post-processing procedure upon the tag sequence. Specifically, we substitute "<SELF>" tokens with input words, remove "<DELETE>" tokens, move tokens that have movements encoded in tags, and, finally, remove spaces between fragments bordered with underscore symbols.

## 3. Experiments

We compare our Thutmose tagger model with Duplex Text Normalization model [3], which shows state-of-the-art results on GTN Dataset and serves as a solid baseline. The Duplex architecture consists of two parts: 1) A tagger that detects the beginning and end of a span that needs to be converted by ITN, and 2) A T5-based [12], [13] decoder that generates the output expression.

### 3.1. Training details

As a backbone for our tagging model, we use pretrained models from HuggingFace library [4]: *bert-base-uncased* and *distilbert-base-uncased* [9] for English, and *DeepPavlov/rubert-base-cased* and *distilbert-base-multilingual-cased* for Russian. We train on 8 V100 16 GB GPU for 6 epochs using batch size 64, optimizer AdamW [14] with 2e-4 learning rate, 0.1 warm up, and 0.1 weight decay.

### 3.2. Evaluation details

The original test data from GTN Dataset contains a single reference for each ITN span. In order to take into account more than one acceptable variant, we prepare a dictionary of multiple possible references. This dictionary is collected automatically from GTN Dataset. It maps the whole input text of ITN to the list of different conversions that occurred with this input anywhere in the corpus. For example, *"the fifteenth of july nineteen forty one"* could have the following valid written forms: *"15 july 1941"*, *"15th july 1941"*, *"15th july, 1941"*, *"15 jul 1941"*. Then, the inferenced span is regarded as correct if all symbols (except spaces) match to at least one possible reference, e.g., "10 sq.ft.", "$10 ft^2$", and "10sq.ft." are valid options for "ten square feet".

In addition to the default test set as in [10], which is a small contiguous part of the full test data in the GTN dataset, we sample another test set (HARD test set in tables 4, 3) - with at least 1000 examples of each semiotic class. This test set is harder because it contains less frequent examples.

We evaluate three metrics. **1. Sentence accuracy** - an automatic metric that matches each prediction with multiple possible variants of the reference. We divide all errors into two groups: "digit error" and "other error". "Digit error" occurs when at least one digit differs from the closest reference variant. The "other error" means a non-digit error is present in the prediction, e.g., punctuation or letter mismatch. **2. Word Error Rate (WER)** - an automatic metric commonly used in ASR. Each prediction is compared with exactly one reference from the initial corpus. **3. Number of unrecoverable errors** shows the number of errors that corrupt the semantics of the input. We manually assess the model's outputs that do not match the reference to estimate this value.

## 4. Results

Table 3 and Table 4 summarize sentence-level accuracy for English and Russian ITN. Sentences with semiotic classes TELEPHONE and ELECTRONIC are removed for comparison as Thutmose Tagger and Duplex model apply different preprocessing which makes results incomparable. Duplex and Thutmose Tagger models show similar results on "digit errors" for English and Russian. At the same time, the Thutmose tagger outperforms Duplex by 1% and 3% sentence accuracy on default and hard Russian test sets, mostly due to "other errors". We observed that Thutmose Tagger works better on cyrillic-to-latin transliterations, while there is no such class of ITN task in English.

Thutmose tagger shows slightly worse WER(+0.8%) on English default test set, while other metrics are better on the same test set. This maybe be partly explained by minor detokenization problems (extra spaces) and by the fact that some translation variants (eg. non-monotonic) are discarded from the corpus during training, but they remain in the test set. The difference between Thutmose tagger with BERT and DistilBERT for English is small: 0.07% and 0.46% sentence accuracy decrease for the default and hard test sets respectively. For Russian this difference is bigger: 0.73% and 2.28% respectively, since multilingual distilBERT is not specifically tuned for Russian language.

Table 3: *Performance metrics (percentage) on English GTN test set, d-BERT stands for distilBERT.*

| Test set | Metric | Duplex model | Thutmose (BERT) | Thutmose (d-BERT) |
|---|---|---|---|---|
| Default | Sent. acc. | 97.31 | **97.43** | 97.36 |
| | Digit error | 0.35 | **0.31** | 0.38 |
| | Other error | 2.34 | **2.26** | **2.26** |
| | WER | **2.9** | 3.7 | 3.74 |
| Hard | Sent. acc. | **85.34** | 85.17 | 84.71 |
| | Digit error | 3.12 | 3.13 | **3.06** |
| | Other error | **11.54** | 11.70 | 12.23 |
| | WER | 9.34 | **9.02** | 9.10 |

---
[4]https://huggingface.co

Table 4: *Performance metrics (percentage) on Russian GTN test set, d-BERT stands for distilBERT.*

| Test set | Metric | Duplex model | Thutmose (BERT) | Thutmose (d-BERT) |
|---|---|---|---|---|
| Default | Sent. acc. | 92.34 | **93.45** | 92.72 |
| | Digit error | 0.51 | **0.43** | 0.52 |
| | Other error | 7.15 | **6.11** | 6.75 |
| | WER | 3.63 | **2.94** | 3.67 |
| Hard | Sent. acc. | 81.02 | **84.03** | 81.75 |
| | Digit error | 3.24 | **3.08** | 3.77 |
| | Other error | 15.74 | **12.90** | 14.48 |
| | WER | 11.76 | **7.07** | 8.05 |

Table 5: *Number of unrecoverable errors for Russian test set.*

| Semiotic class | Total examples | Unrecoverable errors Duplex model | Thutmose tagger |
|---|---|---|---|
| Cardinal | 8352 | 18 (0.22%) | **7 (0.08%)** |
| Ordinal | 1642 | **1 (0.06%)** | 2 (0.12%) |
| Fraction | 1001 | 14 (1.40%) | **12 (1.20%)** |
| Decimal | 1034 | **3 (0.29%)** | 4 (0.39%) |
| Date | 2808 | **0 (0.0%)** | **0 (0.0%)** |
| Measure | 1400 | **2 (0.14%)** | **2 (0.14%)** |
| Money | 1005 | **7 (0.70%)** | 10 (0.99%) |
| Total | 17,242 | 45 (0.26%) | **37 (0.21%)** |

Table 5 shows the number of unrecoverable errors for Russian ITN. The results of the two models are close to each other, except the CARDINAL class on which Duplex model make more mistakes. Note, that Russian ITN task is more difficult than English. For example, ordinals may have different affixes, depending on grammatical context. This makes the tag vocabulary larger and the task more complex. Also there is a large ITN task subclass, when Russian spoken words need to be transliterated to written English (e.g. company names - IBM, Intel).

### 4.1. Error analysis

Table 6 provides examples of Duplex model errors; such errors are typical to seq2seq models. One issue common to seq2seq models is hallucination, which can result in unpredictable and hard to overcome errors. Also, the model might choose to perform normalization with more frequent phrases even when it contradicts the input (first three examples in Table 6). These particular examples are handled correctly by Thutmose tagger.

Table 7 gives examples of Thutmose model errors. They occur sporadically and usually are caused by incorrect alignments in the training corpus. It is easier to analyze the reasons for tagger errors than those of generative models. With the tagger model, we can directly see what tag the model assigns to each token and investigate how similar examples are aligned in the training corpus.

Table 6: *Examples of errors of Duplex model*

| | |
|---|---|
| **Input:** | is about thirty five united states cents |
| **Prediction:** | is about 35 usd |
| **Reference:** | is about 0.35 usd |
| **Input:** | like a twenty dollars table radio for point nine eight |
| **Prediction:** | like a $20 table radio for .99 |
| **Reference:** | like a $20 table radio for .98 |
| **Input:** | air canada seven seven three dropped engine parts on departure |
| **Prediction:** | air canada 777 dropped engine parts on departure |
| **Reference:** | air canada 773 dropped engine parts on departure |
| **Input:** | t o one five six is a phosphodiesterase inhibitor |
| **Prediction:** | tor56 is a phosphodiesterase inhibitor |
| **Reference:** | t- 0156 is a phosphodiesterase inhibitor |

Table 7: *Examples of errors of Thutmose tagger*

| **Input:** | twelve | thousand | seventy | one |
|---|---|---|---|---|
| **Tags:** | _12 | 0 | 07 | 1_ |
| **Prediction:** | 120071 | | | |
| **Reference:** | 12071 | | | |
| **Explanation:** | Duplication due to corpus alignment mistakes. | | | |
| | twelve | thousand | seventy | one |
| | _12 | , | 07 | 1_ (12,071) |
| | _12 | 0 | 7 | 1_ (12071) |
| **Input:** | five | million | croatian | kunas |
| **Tags:** | _5_ | <SELF> | <DEL> | _million__czk_ |
| **Prediction:** | 5 million million czk | | | |
| **Reference:** | 5 million hrk | | | |
| **Explanation:** | 1. tag dictionary misses "hrk" (too rare) | | | |
| 2. Duplication due to corpus alignment mistakes. | | | | |
| | six | million | czech | korunas |
| | _6_ | <DEL> | <DEL> | _million__czk_ |

## 5. Conclusions

We propose a new tagging-based approach to tackle ITN task and apply a dataset construction method based on automatic alignment. Our model, Thutmose, is a single-pass neural model. It does not need separate steps for span detection and decoding, and this simplifies the model pipeline and eliminates span detection errors. One-to-one correspondence in input and output makes it easy to preserve alignment to audio timestamps and to apply custom corrections to some exceptional cases. Thutmose tagger is simpler than the existing approaches, and it achieves state-of-the-art quality results for English and Russian languages on the Google test set. The model performance can be further improved by just improving the alignment algorithms and generating a better training corpus without increasing the model complexity.

## 6. Acknowledgements

The authors would like to thank Yang Zhang and Elena Rastorgueva for their helpful review and feedback.